\documentclass[11pt]{article}

\usepackage[margin=1in]{geometry}
\usepackage[T1]{fontenc}
\usepackage[utf8]{inputenc}
\usepackage{lmodern}
\usepackage{microtype}
\usepackage{amsmath,amssymb}
\usepackage{booktabs}
\usepackage{graphicx}
\usepackage{subcaption}
\usepackage{enumitem}
\usepackage{array}
\usepackage{tabularx}
\usepackage{ragged2e}
\usepackage{placeins}
\usepackage{xcolor}
\usepackage{tikz}
\usepackage{pgfplots}
\usepackage[numbers]{natbib}
\usepackage[colorlinks=true,linkcolor=blue,citecolor=blue,urlcolor=blue]{hyperref}
\usepackage{xspace}

\pgfplotsset{compat=1.18}
\usetikzlibrary{positioning}
\graphicspath{{figures/}}

\newcolumntype{Y}{>{\RaggedRight\arraybackslash}X}
\newcommand{\physiomio}{PhysioMio\xspace}

\hypersetup{
  pdftitle={Deep Neural Networks for Learning Intent from sEMG Signals to Support Hardware Devices for Post-Stroke Neurorehabilitation},
  pdfauthor={Zakariyya Brewster, Divy Wadhwani, Emily Yan, Aidan Wang, Karma Namgyal, Shuting Xie, Markiyan Konyk, Tala Abdelmaguid},
  pdfsubject={Post-stroke five-finger intent decoding from surface electromyography},
  pdfkeywords={surface electromyography, stroke rehabilitation, finger intent, deep learning, knowledge distillation}
}

\title{Deep Neural Networks for Learning Intent from sEMG Signals to Support Hardware Devices for Post-Stroke Neurorehabilitation}
\author{
\small
\begin{tabular}{c@{\hspace{0.7cm}}c}
\begin{tabular}{c}
Zakariyya Brewster\thanks{All authors contributed equally to this work.}\\
Department of Engineering Science\\
University of Toronto
\end{tabular}
&
\begin{tabular}{c}
Divy Wadhwani\\
Computer and Mathematical Sciences\\
University of Toronto Scarborough
\end{tabular}
\\[1.1em]
\begin{tabular}{c}
Emily Yan\\
Department of Computer Science\\
University of Toronto
\end{tabular}
&
\begin{tabular}{c}
Aidan Wang\\
Computer Science and Cognitive Science\\
University of Toronto
\end{tabular}
\\[1.1em]
\begin{tabular}{c}
Karma Namgyal\\
Department of Electrical \& Computer Engineering\\
University of Toronto
\end{tabular}
&
\begin{tabular}{c}
Shuting Xie\\
Department of Computer Science\\
University of Toronto
\end{tabular}
\\[1.1em]
\begin{tabular}{c}
Markiyan Konyk\\
Department of Engineering Science\\
University of Toronto
\end{tabular}
&
\begin{tabular}{c}
Tala Abdelmaguid\\
Department of Computer Engineering\\
University of Toronto
\end{tabular}
\end{tabular}
}
\date{September 2026}

\begin{document}

\maketitle

\begin{abstract}
Finger-specific motor intent is a clinically meaningful control signal for post-stroke neurorehabilitation, where residual muscle activity may remain measurable despite weak or incomplete movement. We study five-finger multilabel intent decoding from impaired-arm high-density surface electromyography (sEMG) in \physiomio, a bilateral longitudinal dataset collected from stroke patients~\citep{physiomioData2026}. A common processing protocol aligns movement labels, applies 20--450 Hz Butterworth filtering and Symlet-4 wavelet denoising, segments overlapping 200 ms windows, and extracts twelve time- and frequency-domain descriptors per channel. Direct LSTM, CNN, and GNN baselines reveal complementary behavior: the LSTM attains the highest subset accuracy (0.545), whereas the GNN attains the highest macro F1 (0.706) and macro AUPRC (0.776). Architecture search then identifies CNN-Large as the strongest single-split CNN, with 0.593 subset accuracy and 0.714 macro F1, while CNN-Micro provides a compact architecture for embedded inference. To match a four-sensor hardware design, we retrain CNN-Micro using channels associated with ECRB, ECRL, FDS, and FDP and exclude the ground electrode from model input. Across five seeds, cross-channel knowledge distillation improves the four-channel student over direct training, reaching $0.5219\pm0.0114$ subset accuracy, $0.7612\pm0.0038$ finger accuracy, and $0.6095\pm0.0058$ macro F1. The selected 123K-parameter model accepts nine windows of 48 features and has been exported to ONNX. These results establish a reproducible software path from post-stroke sEMG to compact five-finger intent prediction for subsequent hardware-in-the-loop evaluation.

\end{abstract}

\section{Introduction}

Stroke frequently disrupts the descending motor pathways needed for coordinated hand opening, grasping, and individuated finger control. For patients with residual forearm muscle activity, surface electromyography (sEMG) offers a non-invasive sensing modality through which intended movement can be estimated even when visible motion is weak or incomplete~\citep{munoznovoa2022upper,zheng2021semg,boukhennoufa2022wearable}. Reliable finger-intent decoding is therefore a central software problem for rehabilitation devices that aim to provide timely assistance, feedback, or task-specific practice.

The decoding problem is difficult because post-stroke sEMG differs substantially from healthy-limb gesture-recognition benchmarks. Paretic recordings are affected by abnormal co-contraction, altered recruitment patterns, fatigue, sensor placement, and impairment severity. In addition, the intended output for rehabilitation is often not a single gesture class but a multi-finger activation pattern suitable for controlling or cueing assistive hardware. A useful model must therefore be evaluated together with the signal-processing and labeling decisions that define the task.

This paper studies that problem as a hardware-aware machine-learning question: how can impaired-arm HD-sEMG be transformed into five-finger intent estimates with models small enough to support responsive Raspberry Pi 5 rehabilitation hardware? The study evaluates the full decoding chain from label alignment and signal processing through model-family comparison, transfer learning, and compact CNN selection. This framing is deliberately tied to embedded rehabilitation use, because a clinically useful decoder must balance prediction quality with model size, latency, and deployability.

Our study is organized around three questions. First, when LSTM, CNN, and GNN decoders are trained on the same \physiomio-derived feature representation, which inductive biases are most useful for multilabel finger-intent prediction? Second, after the baseline comparison, how much can decoding performance be improved through compact CNN architecture search, ResNet teacher references, and healthy-to-impaired transfer learning? Third, how much predictive performance is retained when the high-density representation is reduced to four active channels compatible with the planned sensing hardware, and can cross-channel knowledge distillation narrow that loss?

The central contribution is the formulation and evaluation of post-stroke HD-sEMG finger-intent decoding as a single hardware-aware research problem. We define a reproducible sEMG-to-finger-intent workflow for \physiomio\ recordings, including label alignment, filtering, denoising, windowed feature extraction, patient-level splits, and shared tensor adaptation. We compare LSTM, CNN, and GNN predictors under this common formulation and report both aggregate and finger-wise multilabel metrics. We evaluate an optimized CNN student family against internal ResNet teachers and quantify healthy-to-impaired transfer behavior. Finally, we construct and evaluate a four-active-channel CNN-Micro through direct training, full-to-reduced-channel transfer initialization, and cross-channel knowledge distillation over five random seeds. This last experiment supplies a validated model and explicit input contract for subsequent Raspberry Pi 5 integration while leaving hardware timing and therapeutic evaluation to dedicated studies.

\section{Related Work}

sEMG has long been studied as a non-invasive control channel for rehabilitation, prosthetics, and human-machine interaction, and stroke-specific reviews show that sEMG-driven interventions can support meaningful upper-limb rehabilitation when signal interpretation is reliable enough for assistive use~\citep{munoznovoa2022upper,zheng2021semg}. Broader reviews of wearable sensing for stroke recovery make a related point: clinical value depends on the full stack, including acquisition quality, body placement, labeling protocol, feature representation, and the way machine learning is coupled to assessment or assistive control~\citep{boukhennoufa2022wearable}. For this reason, decoding architecture should not be discussed independently of data construction and deployment context.

Dataset choice is especially important in this area. Ninapro remains the dominant public benchmark family for hand-gesture decoding from sEMG, with ten sub-datasets spanning different electrode layouts, movement vocabularies, and sensor modalities~\citep{ninapro2014,ninaproSite2026}. The widely used DB5 split, for example, records 52 movements from 10 healthy participants with two Myo armbands, while DB1 and DB2 cover larger healthy cohorts with different channel configurations and richer kinematic metadata~\citep{ninapro2014,ninaproSite2026}. By contrast, \physiomio\ was introduced in 2026 as a stroke-specific bilateral and longitudinal HD-sEMG resource with 48 stroke patients, 16 gestures, 64 electrodes, and repeated recordings across inpatient recovery~\citep{physiomioData2026}. That distinction matters because results on healthy-subject multiclass gesture sets are not automatically transferable to impaired-arm intent decoding in post-stroke populations.

Stroke-specific decoding studies have historically been small and heterogeneous. Lee et al.~\citep{lee2010stroke} used subject-specific myoelectric pattern classification for six functional hand movements in chronic stroke survivors and reported a sharp impairment-dependent gap, with mean accuracy of 71.3\% for moderately impaired participants and 37.9\% for severely impaired participants. Anastasiev et al.~\citep{anastasiev2022stroke} later studied post-acute stroke gesture recognition with portable eight-channel sEMG and found that affected-side SVM performance could still reach the high-80\% range on reduced gesture sets, but the task remained sensitive to feature design, label count, and paretic signal quality. More recent deep-learning work has begun to explore richer feature domains and architectures for post-stroke myoelectric recognition; Bao et al.~\citep{bao2024poststroke} evaluated CNN, CNN-LSTM, and CNN-LSTM-attention designs across time, frequency, and wavelet features in chronic stroke participants, with the best reported intra-subject and inter-subject transfer settings reaching 72.95\% and 68.38\% average accuracy, respectively. These studies underscore a recurring challenge for post-stroke intent decoding: strong performance is possible, but dataset scale, subject condition, movement vocabulary, and train-test protocol strongly determine the meaning of any reported number.

Deep neural models broaden that picture but do not remove the comparability problem. Sequence-oriented recurrent models remain natural baselines because they aggregate temporal evidence across windows, while compact 1D CNNs emphasize local temporal motifs and can be compressed aggressively for efficient inference~\citep{hochreiter1997lstm}. Graph neural networks offer a complementary structural view by representing windows as related observations connected through message passing rather than as a purely linear sequence~\citep{kipf2017gcn,hamilton2017graphsage,velickovic2018gat}. On healthy-dataset benchmarks such as Ninapro, deeper end-to-end networks can reach very high multiclass accuracies; for example, Sri-Iesaranusorn et al.~\citep{sri2021dnn} reported 93.87\% accuracy on Ninapro DB5 and 91.69\% on DB7 for 41-movement classification. Other efficient CNN designs on Ninapro DB5 report high-80\% accuracy while explicitly optimizing computation or channel use~\citep{ni2024survey}. Those results are useful context for model capacity, but they address a different population and a different task than impaired-arm multilabel finger-intent prediction.

Transfer learning has recently become one of the most relevant directions for rehabilitation-oriented sEMG. Li et al.~\citep{li2025transfer} showed that transfer-based CNNs can materially improve inter-subject and inter-day robustness on high-density healthy-subject sEMG, and Mohammadiazni et al.~\citep{mohammadiazni2026transfer} demonstrated an especially important stroke-specific result: pretraining on healthy Ninapro DB5 data raised stroke-survivor gesture-classification accuracy from 46\% to 93.6\% in a three-gesture setting. These findings motivate healthy-to-impaired transfer as a natural strategy for paretic-limb decoding, where affected-side training data are comparatively scarce and variable.

Deployment-oriented optimization provides the final piece of context for this manuscript. Knowledge distillation is a standard way to transfer behavior from larger teachers into smaller students~\citep{hinton2015distilling}, and Optuna is widely used for black-box hyperparameter search under practical constraints~\citep{akiba2019optuna}. For rehabilitation wearables, these methods connect offline decoding accuracy to the practical requirements of model size, latency, and hardware-constrained inference.

\begin{table}[!htbp]
  \centering
  \caption{Literature context for the current study. Reported numbers reflect different populations, sensors, gesture vocabularies, and label spaces.}
  \label{tab:literature-context}
  \begingroup
  \scriptsize
  \setlength{\tabcolsep}{3pt}
  \renewcommand{\arraystretch}{1.14}
  \begin{tabularx}{\linewidth}{>{\RaggedRight\arraybackslash}p{2.15cm}YYYY}
\toprule
Source & Population / dataset & Task & Reported performance & Relevance to this paper \\
\midrule
Lee et al. 2010 & 20 chronic stroke survivors; 10 forearm/hand electrodes & Six functional hand movements with subject-specific EMG classification & Mean accuracy 71.3\% for moderate impairment and 37.9\% for severe impairment & Early evidence that stroke severity strongly affects decoder quality \\
Anastasiev et al. 2022 & 19 post-acute stroke patients; portable eight-channel sEMG & Four- to seven-gesture multiclass recognition on affected and non-affected sides & Affected-side SVM accuracy reached 88.7\% on the four-gesture setting & Shows that reduced-vocabulary stroke gesture recognition can reach high accuracy with careful feature design \\
Bao et al. 2024 & 8 chronic stroke participants; post-stroke sEMG feature-domain study & CNN, CNN-LSTM, and CNN-LSTM-attention recognition using time, frequency, and wavelet features & Best reported averages: 72.95\% intra-subject accuracy and 68.38\% inter-subject transfer accuracy & Supports deep architectures for post-stroke decoding while highlighting the difficulty of transfer \\
\parbox[t]{2.15cm}{Mohammadiazni\\et al. 2026} & 10 stroke patients plus healthy Ninapro DB5 pretraining data & Three-gesture classification across multiple arm postures with transfer learning & 93.6\% with transfer learning versus 46\% without transfer & Strong stroke-specific motivation for healthy-to-impaired transfer learning \\
\parbox[t]{2.15cm}{Sri-Iesaranusorn\\et al. 2021} & Ninapro DB5/DB7, mostly healthy-subject settings & 41-movement deep neural network classification & 93.87\% on DB5 and 91.69\% on DB7 & Healthy-data ceiling is high, but the task is easier than impaired-arm multilabel finger decoding \\
This project & \physiomio\ impaired-arm split from 48 stroke patients; patient-level train/val/test split & Five-finger multilabel intent decoding and four-channel hardware-targeted retraining & CNN-Large: 0.593 subset accuracy and 0.714 macro F1; distilled four-channel CNN-Micro: $0.522\pm0.011$ and $0.609\pm0.006$ & Connects impaired-arm model comparison to a concrete reduced-sensor interface \\
\bottomrule
\end{tabularx}

  \endgroup
\end{table}
\FloatBarrier

\section{Method}

\subsection{Overview}

\begin{figure}[t]
  \centering
  \includegraphics[width=\linewidth,trim=35 120 35 70,clip]{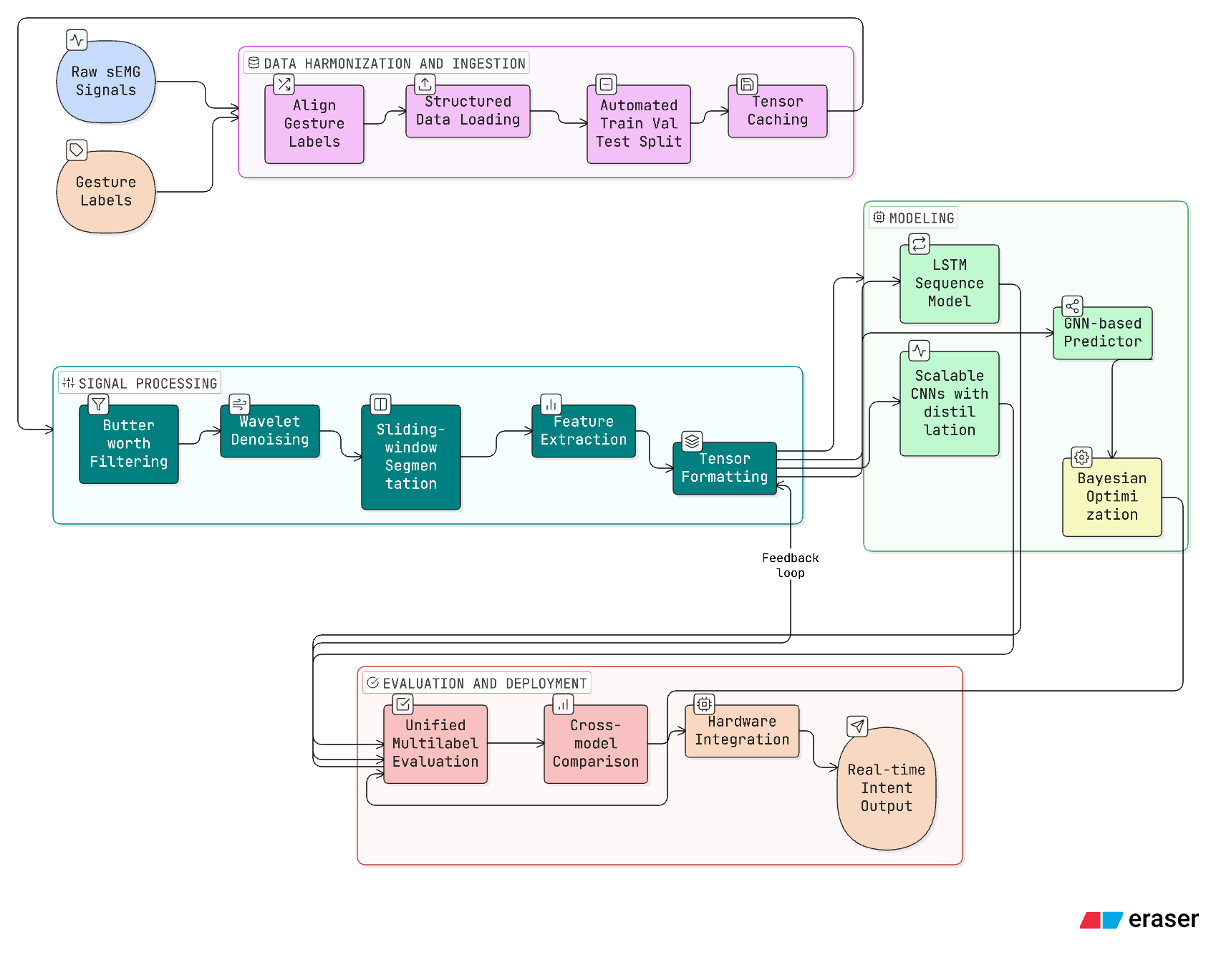}
  \caption{Overview of the sEMG finger-intent decoding pipeline. Raw sEMG and gesture annotations are aligned, segmented, featurized, mapped into shared tensors, evaluated across multiple model families, and prepared for downstream deployment analysis.}
  \label{fig:pipeline}
\end{figure}

This project defines a modular decoding workflow from multichannel sEMG and gesture annotations to five-finger prediction logits. The pipeline contains four stages: data harmonization, signal processing, model inference, and evaluation or deployment analysis. Figure~\ref{fig:pipeline} summarizes the workflow used throughout the study.

For notation, let a pre-segmented raw recording be $X_{\mathrm{raw}} \in \mathbb{R}^{T \times C}$, where $T$ is the number of raw time samples and $C$ is either 64 for the high-density benchmarks or 4 for hardware-targeted retraining. After preprocessing and windowing, the project extracts a feature tensor $X_{\mathrm{feat}} \in \mathbb{R}^{C \times W \times F}$, where $W$ is the number of windows and $F=12$ is the number of handcrafted descriptors per channel-window pair. Each sample is paired with a multilabel target vector $y \in \{0,1\}^5$ indicating intended activation of thumb, index, middle, ring, and little finger.

\subsection{Data harmonization and ingestion}

The implementation builds processed datasets directly from raw \physiomio\ parquet files. The loader searches the raw data tree, groups files by patient, aligns contiguous stretches of the same \texttt{movement\_type}, maps each label to a five-bit finger target, discards contiguous label segments shorter than 200 raw samples, and applies the preprocessing pipeline to each usable segment. At the default 2000 Hz sampling rate, this 200-sample rule is a permissive minimum-duration filter for label segments; feature extraction still requires complete 200 ms windows, corresponding to 400 raw samples at 2000 Hz, because the windowing stage does not zero-pad short segments. This converts gesture-level clinical recordings into segment-level multilabel examples.

The original 64-channel model-family experiments use the project loader's 2000 Hz configuration and select the impaired arm. The later four-channel experiment follows the documented \physiomio\ rate of 2048 Hz. In both paths, patients are partitioned into deterministic 70/10/20 train/validation/test splits, so samples from the same patient do not appear in multiple partitions. Processed samples are padded to a dataset-wide maximum window count for batched tensor operations.

\subsection{Signal preprocessing}

The preprocessing module is explicitly parameterized by a frozen configuration object in the implementation. Raw sEMG first passes through a fourth-order Butterworth band-pass filter with 20--450 Hz cutoffs and zero-phase forward-backward application. This stage suppresses low-frequency motion artifacts and high-frequency electrical noise while preserving time alignment across channels.

The filtered signal is then denoised with wavelet soft-thresholding. The default configuration uses a Symlet-4 basis, four decomposition levels, universal threshold selection, and a median-based noise estimate. The same preprocessing configuration is used for all model families, making downstream differences attributable to the representation and predictor rather than to separate signal-conditioning choices.

After denoising, each continuous segment is divided into 200 ms windows with 50\% overlap and no zero-padding. These settings balance temporal responsiveness against the sample count required for stable summary statistics.

\subsection{Window-level feature representation}

Each channel-window pair is summarized by twelve handcrafted descriptors spanning time and frequency domains. The time-domain set includes root mean square, mean absolute value, integrated EMG, waveform length, variance, zero crossings, slope sign changes, and Willison amplitude. The frequency-domain set includes mean frequency, median frequency, spectral entropy, and total power, with spectral quantities estimated from Welch periodograms using \texttt{nperseg = min(256, len(x))}~\citep{welch1967fft}. For the default 200 ms \physiomio\ window, $\mathrm{len}(x)=400$, so the Welch segment length is 256 samples; shorter windows, if produced by alternate configurations, use their full length. Because no explicit \texttt{noverlap} is passed, SciPy uses its default 50\% periodogram overlap. Degenerate spectra are zeroed explicitly to avoid unstable feature values. The resulting tensor has shape $(C, W, F)$, where $C$ is the number of channels, $W$ is the number of windows, and $F=12$ is the feature count.

This feature representation is a deliberate engineering choice rather than a claim that raw end-to-end learning is unnecessary. Handcrafted time- and frequency-domain summaries reduce the input dimensionality, expose standard sEMG descriptors to all model families, and support compact students intended for embedded inference. They are also appropriate for the current single-dataset setting, where learning stable low-level time-frequency filters directly from raw HD-sEMG would require additional data and ablation experiments.

\subsection{Tensor formatting and shared adapter}

The project standardizes cross-model comparison with one shared adapter. It transforms $(C, W, F)$ or $(N, C, W, F)$ tensors into $(N, W, C \times F)$ sequences, where windows act as time steps and all channel features are concatenated into one vector per step. For a single sample, the adapted sequence can be written as
\[
S = \left[s_1, \ldots, s_W\right]^\top \in \mathbb{R}^{W \times d},
\qquad
s_w = \mathrm{vec}\!\left(X_{\mathrm{feat},:,w,:}\right),
\qquad
d = C F = 768.
\]
This adapter creates a common input space for recurrent, convolutional, and graph-based predictors.

For the CNN branch, the adapted tensor is subsequently permuted to a channels-first layout. The high-density CNN stack treats the $64 \times 12$ channel-feature grid as 768 input channels for temporal \texttt{Conv1d} layers, preserving the same upstream representation while allowing convolutions over window order. The reduced-channel stack analogously uses $4 \times 12=48$ features per window. This is best interpreted as temporal convolution over a feature sequence whose per-window vector already contains spatial-channel and spectral summaries. It is not asserted to be superior to raw-signal 1D CNNs or 2D channel-by-time convolutions; those alternatives remain important ablations for future work.

\subsection{Hardware-targeted channel reduction}

The reduced-input experiment selects four active channels associated with two wrist/finger extensors and two finger flexors in the canonical order [ECRB, ECRL, FDS, FDP]. For the left array, the one-based \physiomio\ channel indices are $[1,3,9,14]$; for the right array they are $[15,16,9,1]$. The corresponding zero-based right-map indices used by the final model are $[14,15,8,0]$. Because the available metadata do not provide a reliable patient-level laterality field for this hardware placement, preprocessing creates left-map and right-map views under an identical patient split. The final experiment uses the right-map view selected during development.

Only four active sEMG signals enter the feature extractor. The hardware ground or reference electrode is not a fifth signal channel and is never included in $X_{\mathrm{raw}}$, $X_{\mathrm{feat}}$, or the neural-network input. At 2048 Hz, a 200 ms window contains 410 samples after rounding and the 50\% overlap produces a 205-sample stride. The final model receives nine consecutive windows, giving an input $S_{4\mathrm{ch}}\in\mathbb{R}^{9\times48}$, and preserves the same five output logits as the high-density models.

\subsection{Predictor families}

The LSTM branch applies two stacked recurrent layers with hidden sizes 128 and 256, followed by a 128-unit fully connected layer and dropout before five output logits. Given adapted sequence input $s_t$, the recurrent dynamics follow the standard LSTM update
\[
\begin{aligned}
i_t &= \sigma(W_i s_t + U_i h_{t-1} + b_i), \\
f_t &= \sigma(W_f s_t + U_f h_{t-1} + b_f), \\
o_t &= \sigma(W_o s_t + U_o h_{t-1} + b_o), \\
\tilde{c}_t &= \tanh(W_c s_t + U_c h_{t-1} + b_c), \\
c_t &= f_t \odot c_{t-1} + i_t \odot \tilde{c}_t, \\
h_t &= o_t \odot \tanh(c_t),
\end{aligned}
\]
and the project uses the final hidden state from the second recurrent layer for classification. This branch emphasizes temporal accumulation across windows and serves as the main sequential baseline for the direct benchmark.

The GNN branch interprets each time window as a node in a graph. The implementation supports GCN, GraphSAGE, and GAT operators~\citep{kipf2017gcn,hamilton2017graphsage,velickovic2018gat}; the benchmarked direct model uses the default two-layer GCN setting with hidden dimension 64, ReLU activations between graph-convolution layers, dropout 0.5, and mean graph pooling. The benchmarked path constructs a complete undirected graph over windows within each sample before graph-level pooling and readout. The resulting edge set has $O(W^2)$ edges per sample, which is computationally feasible for the short padded window sequences used here and lets each window exchange information with all other windows. Sparse temporal adjacency and k-nearest-neighbor graph construction are left as graph-design ablations. At a generic level, the node update can be written as
\[
h_i^{(\ell+1)} =
\psi_\ell\!\left(
h_i^{(\ell)},
\operatorname*{AGG}_{j \in \mathcal{N}(i)}
\phi_\ell\!\left(h_i^{(\ell)}, h_j^{(\ell)}\right)
\right),
\]
followed by graph pooling
\[
g = \operatorname{POOL}\!\left(\{h_i^{(L)}\}_{i=1}^{W}\right)
\]
and a multilabel readout head. In the benchmarked GCN setting, $\operatorname*{AGG}$ is the normalized neighborhood aggregation implemented by \texttt{GCNConv}, $\psi_\ell$ is a linear graph-convolution update followed by ReLU for non-final layers, and \texttt{POOL} is global mean pooling. This branch injects an explicit structural prior: windows are related observations rather than only a flat sequence.

The CNN branch includes five student variants ranging from Nano to XLarge, each built around a $1 \times 1$ projection, temporal convolution blocks, adaptive pooling, and a compact fully connected head. If the channels-first representation is denoted by $\hat{S} \in \mathbb{R}^{d \times W}$, the student stack can be summarized as
\[
H^{(0)} = \rho\!\left(\operatorname{BN}(W_p * \hat{S} + b_p)\right),
\qquad
H^{(\ell+1)} = \rho\!\left(\operatorname{BN}(W_\ell * H^{(\ell)} + b_\ell)\right),
\]
where $*$ denotes 1D convolution over window order and $\rho$ is a ReLU nonlinearity. Adaptive pooling compresses the temporal axis before a multilayer perceptron produces the five logits. The student family spans approximately 54K parameters and 53 KB INT8 for Nano to 1.62M parameters and 1.58 MB INT8 for XLarge. The teacher family uses 1D ResNet-50, ResNet-101, and ResNet-152 variants with bottleneck residual blocks.

\subsection{Training, tuning, and evaluation protocol}

The training utilities unify optimization and reporting across model families. The current implementation uses binary cross-entropy with logits for five-label prediction, Adam-based optimization, plateau-triggered learning-rate reduction, early stopping, checkpointing, and saved metric curves. For logits $z \in \mathbb{R}^5$, the base multilabel objective is
\[
\mathcal{L}_{\mathrm{BCE}}
=
-\frac{1}{5}
\sum_{k=1}^{5}
\left[
y_k \log \sigma(z_k)
+ (1-y_k) \log \left(1-\sigma(z_k)\right)
\right].
\]
Evaluation reports subset accuracy, also called exact match ratio, finger-level averaged accuracy, macro precision, macro recall, macro F1, macro AUROC, and macro AUPRC. Subset accuracy counts a sample as correct only when all five finger labels are predicted correctly. The metrics module also stores per-finger scores and task-level curve data, which enables both aggregate and finger-wise analysis after training.

For the CNN students, Optuna searches architecture and training choices for each model size~\citep{akiba2019optuna}. For a trial with validation macro F1 $F_{\mathrm{val}}$ reported on a 0--100 scale, measured latency $t$, and size-specific baseline latency $t_{\mathrm{ref}}$ measured after one baseline epoch, the objective is
\[
\operatorname{score}
= \frac{F_{\mathrm{val}}}{100}\, p\!\left(\frac{t}{t_{\mathrm{ref}}}\right),
\]
where
\[
p(r)=
\begin{cases}
1, & r \le 1,\\
1 - 3(r-1), & 1 < r \le 1.2,\\
0.05, & r > 1.2.
\end{cases}
\]
Thus trials at or below the baseline latency are not penalized, mildly slower trials receive a linear penalty, and substantially slower trials are strongly downweighted. Trial-level early stopping and median pruning are used during search.

The system also implements teacher-ensemble distillation, per-finger threshold tuning, and latency benchmarking. For multilabel distillation, the implemented loss blends the hard target term with a temperature-scaled soft target from the teacher,
\[
\mathcal{L}_{\mathrm{KD}}
=
\alpha \, \mathcal{L}_{\mathrm{BCE}}(z_s, y)
+
(1-\alpha) T^2
\operatorname{BCELogits}\!\left(\frac{z_s}{T}, \sigma\!\left(\frac{z_t}{T}\right)\right),
\]
where $z_s$ and $z_t$ are student and teacher logits, $T$ is the distillation temperature, and $\alpha$ controls the hard-soft balance. In the four-channel experiment, paired examples present the full 768-feature representation to the fixed teacher and the matching 48-feature representation to the student. Training uses $T=2$ and $\alpha=0.5$. Per-finger thresholds are selected on validation probabilities and then applied once to test predictions. A separate latency utility can report mean, median, and 95th-percentile inference time on a chosen device, but no Raspberry Pi timing is claimed in this study.

\subsection{Transfer-learning paths}

In addition to direct impaired-arm benchmarks, the study evaluates two distinct transfer mechanisms. In the first, CNN and LSTM models are pretrained on healthy-arm recordings and finetuned on impaired-arm recordings while preserving the same five-finger multilabel objective. The reported CNN experiment uses the CNN-Base configuration selected by the two-stage tuning run, so its rows do not duplicate either the legacy direct CNN baseline or the optimized CNN student sweep.

The four-channel study instead transfers across input density. A compatible 64-channel CNN-Micro source model is trained on the same patient split. Its first projection-layer weights are sliced to the 48 feature indices corresponding to the selected four channels, while shape-compatible later layers are copied directly. This initialization tests whether a full-grid representation transfers to the hardware-feasible sensor subset; it is evaluated separately from cross-channel distillation.

\section{Experiments}

\subsection{Dataset and split construction}

All experiments use the \physiomio\ processing path implemented for this study~\citep{physiomioData2026,psrrepo2026}. Dataset construction extracts contiguous movement segments from raw parquet recordings, maps gesture labels to five-bit finger targets, applies the signal-processing and feature-extraction stages, and stores tensors as PyTorch split files.

Patients are shuffled with a fixed seed and divided into 70\% train, 10\% validation, and 20\% test partitions. The main benchmarks use impaired-arm recordings. The healthy-to-impaired experiments use healthy-arm recordings for pretraining and impaired-arm recordings for finetuning. The four-channel study holds the patient assignment fixed across all training modes and contains 34 training patients (2656 samples), five validation patients (432 samples), and nine test patients (720 samples).

\subsection{Experimental groups}

The first group compares direct LSTM, CNN, and GNN baselines trained under the same 64-channel preprocessing and feature representation. The second evaluates optimized CNN students and ResNet teachers, emphasizing multilabel performance and model footprint. The third evaluates healthy-to-impaired transfer for CNN-Base and LSTM.

The fourth group targets the four-active-channel hardware constraint. Development runs compare left, right, and dual channel-map views and context lengths corresponding to one, four, and nine windows. The selected right-map, nine-window configuration is then trained under three modes: direct supervision, first-layer-sliced transfer from a compatible 64-channel CNN-Micro, and cross-channel distillation from a full-representation teacher. Each final mode is run with seeds 0--4. A separate 64-channel CNN-Micro source run with seed 42 provides a same-split reference but is not included in the five-seed variance estimate.

\subsection{Evaluation and model selection}

The task is five-label multilabel classification. We report subset accuracy (exact match ratio), mean finger accuracy, macro precision, macro recall, macro F1, macro AUROC, macro AUPRC, and per-finger metrics. Subset accuracy counts a sample as correct only when all five finger labels are correct. The baseline, CNN-sweep, ResNet, and healthy-transfer tables contain single deterministic-split estimates and therefore do not support statistical-significance claims.

For the four-channel final runs, per-finger thresholds are tuned on validation probabilities and applied to the held-out test set. Results are reported as the mean and sample standard deviation across five seeds; the artifact bundle additionally records 95\% confidence-interval half-widths computed as $1.96s/\sqrt{5}$. The training mode is chosen by mean validation macro F1, and the handoff checkpoint is the seed within that mode with the highest validation macro F1. Test performance is not used to choose the checkpoint.

\section{Results}

\subsection{Direct baseline models: LSTM, CNN, and GNN}

Table~\ref{tab:model-comparison} reports the direct comparison among LSTM, CNN, and GNN decoders on the impaired-arm test split. The three families express different assumptions about the same windowed feature sequence. The LSTM treats the input as an ordered temporal process, the CNN searches for local temporal motifs after channel-feature projection, and the GNN treats windows as nodes in a sample-level graph. Their results show complementary behavior: the LSTM leads subset accuracy and macro AUROC, while the GNN leads macro F1 and macro AUPRC.

\begin{table}[!htbp]
  \centering
  \small
  \caption{Held-out test performance for the three direct model families. Subset accuracy is the multilabel exact match ratio. Bold numbers mark the best score in each column within this baseline comparison.}
  \label{tab:model-comparison}
  \begin{tabular}{lccccc}
\toprule
Model & Subset Acc. & Finger Acc. & Macro F1 & Macro AUROC & Macro AUPRC \\
\midrule
LSTM & \textbf{0.545} & \textbf{0.784} & 0.705 & \textbf{0.858} & 0.754 \\
CNN legacy & 0.442 & 0.694 & 0.676 & 0.767 & 0.764 \\
GNN  & 0.448 & 0.683 & \textbf{0.706} & 0.787 & \textbf{0.776} \\
\bottomrule
\end{tabular}

\end{table}

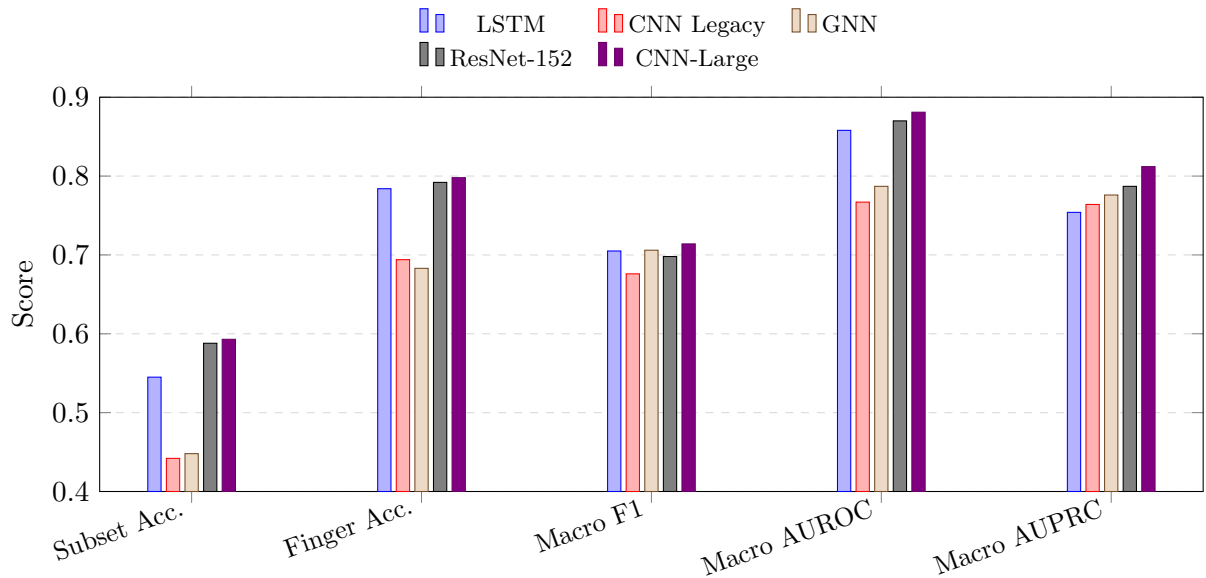
\begin{figure}[!htbp]
  \centering
  \begin{tikzpicture}
\begin{axis}[
  ybar,
  width=0.98\linewidth,
  height=6.8cm,
  ymin=0.40,
  ymax=0.90,
  ylabel={Score},
  symbolic x coords={Subset Acc.,Finger Acc.,Macro F1,Macro AUROC,Macro AUPRC},
  xtick=data,
  xticklabel style={rotate=20,anchor=east,font=\small},
  ymajorgrids=true,
  grid style={dashed,gray!30},
  legend style={
    at={(0.5,1.04)},
    anchor=south,
    legend columns=3,
    draw=none,
    font=\footnotesize,
    /tikz/every even column/.append style={column sep=0.7em}
  },
  bar width=5pt,
]
\addplot coordinates {(Subset Acc.,0.545) (Finger Acc.,0.784) (Macro F1,0.705) (Macro AUROC,0.858) (Macro AUPRC,0.754)};
\addplot coordinates {(Subset Acc.,0.442) (Finger Acc.,0.694) (Macro F1,0.676) (Macro AUROC,0.767) (Macro AUPRC,0.764)};
\addplot coordinates {(Subset Acc.,0.448) (Finger Acc.,0.683) (Macro F1,0.706) (Macro AUROC,0.787) (Macro AUPRC,0.776)};
\addplot coordinates {(Subset Acc.,0.588) (Finger Acc.,0.792) (Macro F1,0.698) (Macro AUROC,0.870) (Macro AUPRC,0.787)};
\addplot coordinates {(Subset Acc.,0.593) (Finger Acc.,0.798) (Macro F1,0.714) (Macro AUROC,0.881) (Macro AUPRC,0.812)};
\legend{LSTM,CNN Legacy,GNN,ResNet-152,CNN-Large}
\end{axis}
\end{tikzpicture}
  \caption{Aggregate-metric comparison across the direct baselines and the strongest internal teacher and student references from the optimized CNN branch.}
  \label{fig:model-comparison}
\end{figure}

The baseline comparison suggests that temporal recurrence and graph aggregation capture different aspects of paretic sEMG. The LSTM's stronger subset accuracy indicates better all-label consistency, whereas the GNN's macro F1 and AUPRC indicate stronger class-balanced discrimination despite lower finger accuracy. The legacy CNN is weaker in this stage, motivating a systematic convolutional architecture search.
\FloatBarrier

\subsection{Optimized CNN students and ResNet teachers}

Table~\ref{tab:cnn-sweep} evaluates the optimized CNN student family together with internal ResNet teacher models. CNN-Large gives the strongest single-split aggregate CNN result, reaching 0.593 subset accuracy, 0.798 finger accuracy, 0.714 macro F1, 0.881 macro AUROC, and 0.812 macro AUPRC. The 158K-parameter CNN-Micro retains 0.578 subset accuracy and 0.697 macro F1 at an estimated 154 KB INT8 weight footprint, motivating its selection as the compact architecture for subsequent hardware-targeted study.

\begin{table}[!htbp]
  \centering
  \small
  \caption{CNN-family evaluation. Student parameter counts and INT8 sizes describe model scale; ResNet rows provide high-capacity teacher references. Bold numbers mark the best score in each column, and $\dagger$ marks the compact architecture carried into deployment-oriented development.}
  \label{tab:cnn-sweep}
  \resizebox{\linewidth}{!}{\begin{tabular}{llccccccc}
\toprule
Variant & Type & Params & INT8 Size & Subset Acc. & Finger Acc. & Macro F1 & Macro AUROC & Macro AUPRC \\
\midrule
ResNet-50  & Teacher & 16.3M & 15.6 MB & 0.570 & 0.788 & 0.656 & 0.840 & 0.772 \\
ResNet-101 & Teacher & 28.6M & 27.3 MB & 0.569 & 0.795 & 0.681 & 0.852 & 0.767 \\
ResNet-152 & Teacher & 38.8M & 36.9 MB & 0.588 & 0.792 & 0.698 & 0.870 & 0.787 \\
\midrule
Nano        & Student & 54K   & 53 KB   & 0.580 & 0.794 & 0.698 & 0.855 & 0.758 \\
Micro$^\dagger$ & Student & 158K  & 154 KB  & 0.578 & 0.794 & 0.697 & 0.871 & 0.793 \\
Base        & Student & 390K  & 381 KB  & 0.575 & \textbf{0.798} & 0.707 & 0.870 & 0.775 \\
Large       & Student & 803K  & 784 KB  & \textbf{0.593} & \textbf{0.798} & \textbf{0.714} & \textbf{0.881} & \textbf{0.812} \\
XLarge      & Student & 1.62M & 1.58 MB & 0.551 & 0.794 & 0.690 & 0.830 & 0.726 \\
\bottomrule
\end{tabular}
}
\end{table}

CNN-Micro is within 0.015 subset accuracy and 0.017 macro F1 of CNN-Large while using approximately one fifth of its estimated INT8 footprint. The sweep is not monotonic in parameter count: CNN-XLarge is the largest student but performs below CNN-Large on every reported metric. This pattern is consistent with an intermediate capacity being better matched to the available dataset and search budget; confirming overfitting would require learning-curve and regularization ablations. ResNet-152 is the strongest teacher, with 0.588 subset accuracy and 0.698 macro F1, while CNN-Large has higher single-split values on the principal aggregate metrics.
\FloatBarrier

\subsection{Healthy-to-impaired transfer}

Table~\ref{tab:transfer-learning} summarizes healthy-to-impaired transfer learning. This experiment uses the tuned CNN-Base two-stage configuration and is separate from both the legacy direct CNN and optimized student sweep. CNN-Base finetuning improves subset accuracy from 0.465 to 0.585, finger accuracy from 0.760 to 0.800, macro AUROC from 0.833 to 0.863, and macro AUPRC from 0.699 to 0.767. The LSTM behaves differently: subset and finger accuracy increase after finetuning, but macro F1, AUROC, and AUPRC decrease.

\begin{table}[!htbp]
  \centering
  \small
  \caption{Healthy-to-impaired transfer-learning summaries. CNN rows use the tuned CNN-Base configuration. Bold values compare pretraining and finetuning within each architecture.}
  \label{tab:transfer-learning}
  \begin{tabular}{lccccc}
\toprule
Model / Stage & Subset Acc. & Finger Acc. & Macro F1 & Macro AUROC & Macro AUPRC \\
\midrule
CNN-Base pre. & 0.465 & 0.760 & 0.664 & 0.833 & 0.699 \\
CNN-Base fine. & \textbf{0.585} & \textbf{0.800} & \textbf{0.680} & \textbf{0.863} & \textbf{0.767} \\
LSTM pre. & 0.523 & 0.752 & \textbf{0.679} & \textbf{0.858} & \textbf{0.773} \\
LSTM fine. & \textbf{0.574} & \textbf{0.777} & 0.660 & 0.830 & 0.736 \\
\bottomrule
\end{tabular}

\end{table}

These results support healthy-to-impaired transfer for the CNN-Base configuration, but they do not establish transfer as uniformly beneficial across architectures or metrics.
\FloatBarrier

\subsection{Four-channel hardware-targeted retraining}

Table~\ref{tab:four-channel} reports the final CNN-Micro study under the four-active-channel input constraint. Direct training reaches $0.5822\pm0.0162$ macro F1. First-layer-sliced transfer does not improve this result, attaining $0.5706\pm0.0109$. Cross-channel distillation gives the strongest mean performance among the reduced-input modes, with $0.5219\pm0.0114$ subset accuracy, $0.7612\pm0.0038$ finger accuracy, $0.6095\pm0.0058$ macro F1, $0.7904\pm0.0073$ macro AUROC, and $0.6933\pm0.0036$ macro AUPRC.

\begin{table}[!htbp]
  \centering
  \small
  \caption{Hardware-targeted CNN-Micro results. Four-channel entries are mean $\pm$ sample standard deviation across five seeds. The 64-channel source is a separate single-seed reference and is not included in the variance comparison. Bold values mark the best four-channel mean.}
  \label{tab:four-channel}
  \resizebox{\linewidth}{!}{\begin{tabular}{lcccccc}
\toprule
Training mode & Seeds & Subset Acc. & Finger Acc. & Macro F1 & Macro AUROC & Macro AUPRC \\
\midrule
64-channel source & 1 & 0.5658 & 0.7846 & 0.6810 & 0.8520 & 0.7696 \\
Direct 4-channel & 5 & $0.4801\pm0.0080$ & $0.7326\pm0.0127$ & $0.5822\pm0.0162$ & $0.7683\pm0.0139$ & $0.6401\pm0.0230$ \\
Transfer 4-channel & 5 & $0.4618\pm0.0089$ & $0.7111\pm0.0078$ & $0.5706\pm0.0109$ & $0.7478\pm0.0064$ & $0.5985\pm0.0164$ \\
Distilled 4-channel & 5 & $\mathbf{0.5219\pm0.0114}$ & $\mathbf{0.7612\pm0.0038}$ & $\mathbf{0.6095\pm0.0058}$ & $\mathbf{0.7904\pm0.0073}$ & $\mathbf{0.6933\pm0.0036}$ \\
\bottomrule
\end{tabular}
}
\end{table}

\begin{figure}[!htbp]
  \centering
  \begin{tikzpicture}
\begin{axis}[
  ybar,
  width=0.90\linewidth,
  height=6.2cm,
  ymin=0.44,
  ymax=0.80,
  ylabel={Mean score},
  symbolic x coords={Subset Acc.,Finger Acc.,Macro F1,Macro AUROC,Macro AUPRC},
  xtick=data,
  xticklabel style={rotate=18,anchor=east,font=\small},
  ymajorgrids=true,
  grid style={dashed,gray!30},
  legend style={
    at={(0.5,1.03)},
    anchor=south,
    legend columns=3,
    draw=none,
    font=\footnotesize,
    /tikz/every even column/.append style={column sep=0.8em}
  },
  bar width=8pt,
]
\addplot coordinates {(Subset Acc.,0.4801) (Finger Acc.,0.7326) (Macro F1,0.5822) (Macro AUROC,0.7683) (Macro AUPRC,0.6401)};
\addplot coordinates {(Subset Acc.,0.4618) (Finger Acc.,0.7111) (Macro F1,0.5706) (Macro AUROC,0.7478) (Macro AUPRC,0.5985)};
\addplot coordinates {(Subset Acc.,0.5219) (Finger Acc.,0.7612) (Macro F1,0.6095) (Macro AUROC,0.7904) (Macro AUPRC,0.6933)};
\legend{Direct,Transfer,Distilled}
\end{axis}
\end{tikzpicture}
  \caption{Mean performance of the three four-channel CNN-Micro training modes. Distillation improves all displayed aggregate metrics relative to direct training and transfer initialization.}
  \label{fig:four-channel}
\end{figure}
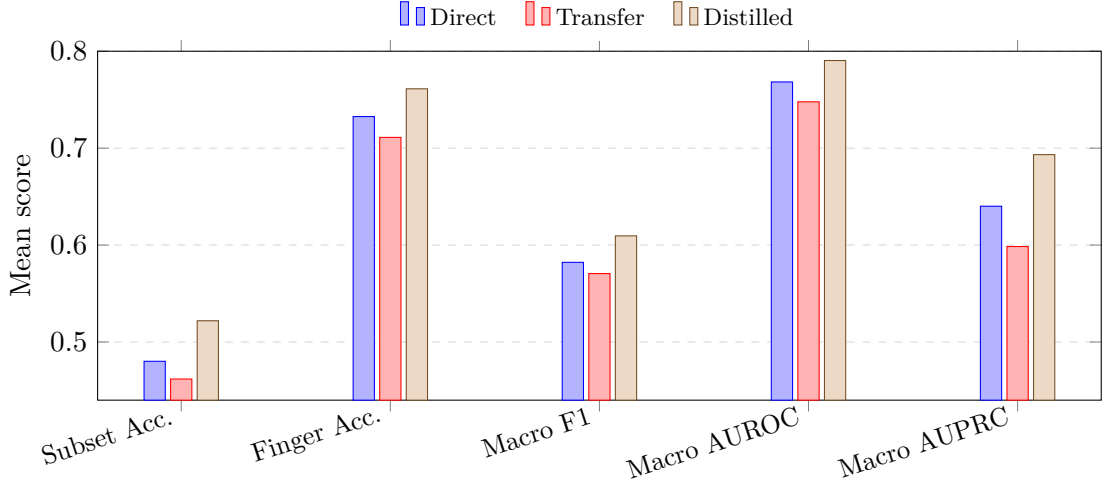

Relative to direct four-channel training, distillation increases mean subset accuracy by 0.0418, finger accuracy by 0.0287, and macro F1 by 0.0272. The separate 64-channel source remains stronger, with 0.6810 macro F1 and 0.5658 subset accuracy, quantifying the cost of reducing the sensor grid. The selected four-channel network has 123,317 parameters, accepts a fixed $(N,9,48)$ tensor, and produces five logits. The seed-4 checkpoint was selected by validation macro F1 and exported to ONNX; no on-device timing result is reported.
\FloatBarrier

\subsection{Finger-level behavior and external positioning}

Finger-wise scores in the original model-family study show that ranking varies across outputs. Table~\ref{tab:per-finger-f1} compares the LSTM and GNN direct baselines with CNN-Large. The GNN is strongest on thumb, CNN-Large leads on index and little finger, and the LSTM is slightly strongest on middle and ring finger. These patterns are plausible in light of hand biomechanics but remain hypotheses rather than anatomical evidence. Thumb activation is more anatomically distinct from the ulnar fingers, whereas middle and ring fingers frequently participate in coupled grasping synergies. Architecture-specific per-finger behavior therefore deserves targeted validation before control policies prioritize particular outputs.

\begin{table}[!htbp]
  \centering
  \small
  \caption{Per-finger F1 score in the original model-family study. Bold values mark the strongest model for each finger.}
  \label{tab:per-finger-f1}
  \begin{tabular}{lccc}
\toprule
Finger & LSTM & GNN & CNN-Large \\
\midrule
Thumb  & 0.716 & \textbf{0.776} & 0.729 \\
Index  & 0.723 & 0.696 & \textbf{0.750} \\
Middle & \textbf{0.712} & 0.689 & 0.698 \\
Ring   & \textbf{0.696} & 0.688 & 0.695 \\
Little & 0.680 & 0.678 & \textbf{0.696} \\
\bottomrule
\end{tabular}

\end{table}

\begin{figure}[!htbp]
  \centering
  \begin{tikzpicture}
\begin{axis}[
  ybar,
  width=0.98\linewidth,
  height=6.8cm,
  ymin=0.60,
  ymax=0.80,
  ylabel={Per-finger F1},
  symbolic x coords={Thumb,Index,Middle,Ring,Little},
  xtick=data,
  ymajorgrids=true,
  grid style={dashed,gray!30},
  legend style={
    at={(0.5,1.04)},
    anchor=south,
    legend columns=3,
    draw=none,
    font=\footnotesize,
    /tikz/every even column/.append style={column sep=0.8em}
  },
  bar width=8pt,
]
\addplot coordinates {(Thumb,0.716) (Index,0.723) (Middle,0.712) (Ring,0.696) (Little,0.680)};
\addplot coordinates {(Thumb,0.776) (Index,0.696) (Middle,0.689) (Ring,0.688) (Little,0.678)};
\addplot coordinates {(Thumb,0.729) (Index,0.750) (Middle,0.698) (Ring,0.695) (Little,0.696)};
\legend{LSTM,GNN,CNN-Large}
\end{axis}
\end{tikzpicture}
  \caption{Per-finger F1 comparison derived from committed metric files. Aggregate CNN gains coexist with output-specific strengths for the LSTM and GNN baselines.}
  \label{fig:per-finger-f1}
\end{figure}
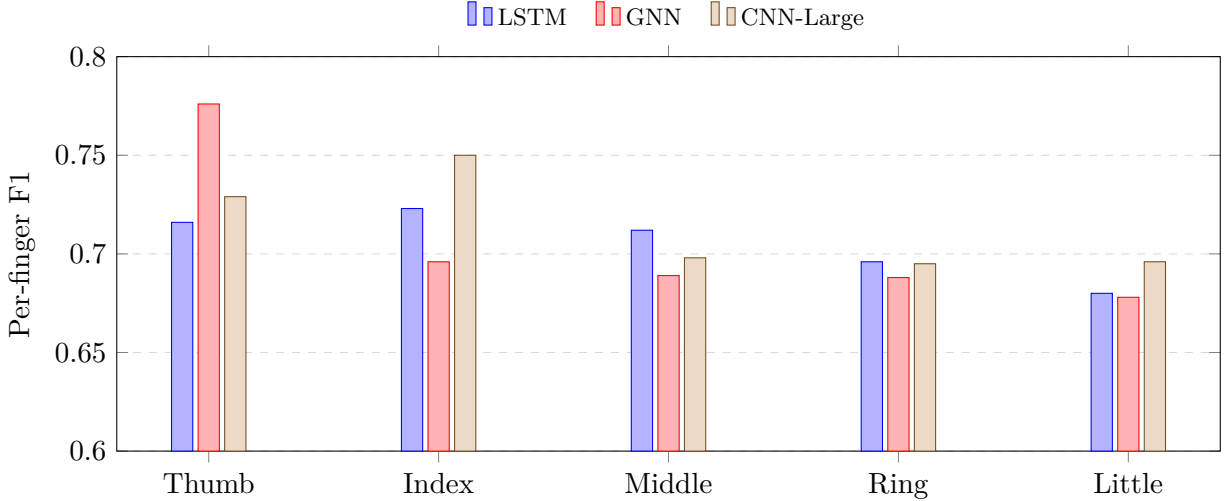

Table~\ref{tab:literature-context} places these results beside prior stroke and sEMG studies. The comparison is not a leaderboard because populations, sensors, labels, and split protocols differ. The present task uses impaired-arm \physiomio\ recordings, patient-level splits, and a five-label subset-accuracy metric requiring all finger activations to be correct simultaneously. Within that setting, the optimized CNN branch is competitive with internal ResNet teachers, and the reduced-channel study demonstrates the measurable trade-off between sensing density and an implementable input contract.

\section{Discussion}

The direct comparison shows that the five-finger task benefits from more than one representation of temporal structure. The LSTM's subset-accuracy advantage indicates greater consistency across the complete output vector, whereas the GNN's macro F1 and AUPRC indicate stronger class-balanced discrimination. These complementary results are consistent with paretic sEMG containing both ordered temporal dynamics and relationships among non-adjacent windows.

Architecture search changes the interpretation of the initial CNN baseline. The tuned CNN students outperform the legacy CNN on every aggregate metric, and CNN-Large gives the strongest offline result in that family. The non-monotonic decline for CNN-XLarge indicates that capacity alone does not determine performance under the available data and search budget. CNN-Micro retains most of CNN-Large's performance at a substantially smaller estimated footprint, providing the architectural basis for reduced-channel development.

The two transfer-learning studies produce different outcomes. Healthy-to-impaired staging improves the CNN-Base configuration but gives mixed LSTM results, consistent with transfer being architecture- and metric-dependent. In the reduced-channel study, directly slicing the compatible 64-channel CNN-Micro initialization does not outperform training from scratch. The selected four channels may not preserve the spatial basis encoded by the full first-layer weights, so reusing those weights constrains the student without supplying all of the source information.

Cross-channel distillation is more effective because it transfers output behavior rather than requiring direct correspondence between full-grid and reduced-grid filters. Across five seeds, the distilled four-channel student improves mean subset accuracy, finger accuracy, macro F1, macro AUROC, and macro AUPRC relative to direct training. Its lower variance on macro F1 and ranking metrics also suggests more consistent optimization under the tested seeds. Nevertheless, the 64-channel source remains stronger, and the 0.0715 macro-F1 difference quantifies information lost when the input is reduced to four sensors. This is a meaningful engineering trade-off: the four-channel model sacrifices some offline discrimination in exchange for a sensor contract that can be implemented by the planned device.

The work advances post-stroke sEMG decoding in three connected ways. It evaluates recurrent, convolutional, and graph models under one impaired-arm multilabel formulation; measures model-development strategies rather than assuming that transfer or scaling will help; and links the final learning problem to a concrete four-sensor input. Comparisons with Ninapro and reduced-vocabulary stroke studies must account for differences in participants, labels, sensors, and split protocols, but the internal comparison against 1D ResNet teachers shows that compact temporal CNNs are credible for this \physiomio\ task. The contribution is therefore both empirical and systems-oriented: a reproducible analysis of model choice followed by a measured transition from high-density recordings to deployable sensing.

The selected 123K-parameter student has been exported to a fixed-context ONNX graph, completing the software path from raw signal chunks to five thresholded finger predictions. This establishes interface compatibility, not real-time hardware performance. The current preprocessing reruns zero-phase filtering and wavelet denoising over rolling history, and its latency must be measured on the Raspberry Pi 5 before timing guarantees can be made.

Several limitations define the next experiments. The original model-family, architecture-search, ResNet, and healthy-transfer results are single-run estimates. The four-channel comparison adds five random seeds but retains one patient split; patient-level cross-validation is still needed to characterize generalization uncertainty. Electrode maps were inferred from the intended hardware placement rather than validated with signals collected by the final sensor assembly. Raw-signal CNNs, alternative channel-selection methods, sparse GNN graphs, and causal preprocessing remain untested ablations. Finally, the reported endpoints measure decoding rather than therapeutic benefit. Hardware-in-the-loop timing, robustness to electrode shift and signal drift, and studies with the target rehabilitation population are required before the model can support clinical conclusions.

\section{Conclusion}

This study developed a post-stroke five-finger intent decoder from impaired-arm sEMG and evaluated its progression from model-family comparison to hardware-targeted channel reduction. LSTM and GNN baselines expose complementary temporal behavior, while CNN architecture search identifies a compact convolutional path. Healthy-to-impaired transfer helps the CNN-Base experiment but not every architecture or metric.

For the four-active-channel design, cross-channel distillation produces the strongest reduced-input CNN-Micro, reaching $0.6095\pm0.0058$ macro F1 and $0.5219\pm0.0114$ subset accuracy across five seeds. The resulting 123K-parameter model accepts nine windows of ECRB, ECRL, FDS, and FDP features and emits five finger-intent logits through a verified ONNX interface. The central engineering result is that useful post-stroke finger-intent information can be retained under a practical four-sensor constraint, providing a concrete software foundation for Raspberry Pi 5 timing tests and subsequent hardware-in-the-loop rehabilitation research.

\bibliographystyle{plainnat}
\bibliography{references}

@misc{psrrepo2026,
  author = {Brewster, Zakariyya and Wadhwani, Divy and Yan, Emily and Wang, Aidan and Namgyal, Karma and Xie, Shuting and Konyk, Markiyan and Abdelmaguid, Tala},
  title = {{sEMG Finger-Intent Decoding Software for Post-Stroke Neurorehabilitation}},
  year = {2026},
  howpublished = {\url{https://github.com/post-stroke-rehab/psr-pipeline}},
  note = {Project software implementation, accessed July 28, 2026}
}

@article{physiomioData2026,
  title = {PhysioMio: Bilateral and Longitudinal HD-sEMG Dataset of 16 Hand Gestures from 48 Stroke Patients},
  author = {Ilg, Julian and Oldemeier, Alexander C. R. and Fieweger, Marie and Deuschel, Luca and Rieckmann, Peter and Young, Peter and Krause, Sabine and Lueth, Tim C.},
  journal = {Scientific Data},
  volume = {13},
  number = {1},
  pages = {19},
  year = {2026},
  doi = {10.1038/s41597-026-06557-0}
}

@misc{ninaproSite2026,
  author = {{Ninapro Project}},
  title = {{The Non-Invasive Adaptive Prosthetics (Ninapro) Database}},
  year = {2026},
  howpublished = {\url{https://ninapro.hevs.ch/}},
  note = {Project site, accessed July 28, 2026}
}

@article{ninapro2014,
  title = {Electromyography Data for Non-Invasive Naturally-Controlled Robotic Hand Prostheses},
  author = {Atzori, Manfredo and Gijsberts, Arjan and Castellini, Claudio and Caputo, Barbara and Hager, Anne-Gabrielle Mittaz and Elsig, Simone and Giatsidis, Giorgio and Bassetto, Franco and M{\"u}ller, Henning},
  journal = {Scientific Data},
  volume = {1},
  pages = {140053},
  year = {2014},
  doi = {10.1038/sdata.2014.53}
}

@article{munoznovoa2022upper,
  title = {Upper Limb Stroke Rehabilitation Using Surface Electromyography: A Systematic Review and Meta-Analysis},
  author = {Munoz-Novoa, Maria and Kristoffersen, Morten B. and Sunnerhagen, Katharina S. and Naber, Autumn and Alt Murphy, Margit and Ortiz-Catalan, Max},
  journal = {Frontiers in Human Neuroscience},
  volume = {16},
  pages = {897870},
  year = {2022},
  doi = {10.3389/fnhum.2022.897870}
}

@article{boukhennoufa2022wearable,
  title = {Wearable Sensors and Machine Learning in Post-Stroke Rehabilitation Assessment: A Systematic Review},
  author = {Boukhennoufa, Issam and Zhai, Xiaojun and Utti, Victor and Jackson, Jo and McDonald-Maier, Klaus},
  journal = {Biomedical Signal Processing and Control},
  volume = {71},
  pages = {103197},
  year = {2022},
  doi = {10.1016/j.bspc.2021.103197}
}

@article{ni2024survey,
  title = {A Survey on Hand Gesture Recognition Based on Surface Electromyography: Fundamentals, Methods, Applications, Challenges and Future Trends},
  author = {Ni, Sike and Al-qaness, Mohammed A. A. and Hawbani, Ammar and Al-Alimi, Dalal and Abd Elaziz, Mohamed E. and Ewees, Ahmed A.},
  journal = {Applied Soft Computing},
  volume = {166},
  pages = {112235},
  year = {2024},
  doi = {10.1016/j.asoc.2024.112235}
}

@article{zheng2021semg,
  title = {Surface Electromyography as a Natural Human-Machine Interface: A Review},
  author = {Zheng, Mingde and Crouch, Michael S. and Eggleston, Michael S.},
  journal = {arXiv preprint arXiv:2101.04658},
  year = {2021}
}

@article{lee2010stroke,
  title = {Subject-Specific Myoelectric Pattern Classification of Functional Hand Movements for Stroke Survivors},
  author = {Lee, Sang Wook and Wilson, Kristin and Lock, Blair A. and Kamper, Derek G.},
  journal = {IEEE Transactions on Neural Systems and Rehabilitation Engineering},
  volume = {19},
  number = {5},
  pages = {558--566},
  year = {2010},
  doi = {10.1109/TNSRE.2010.2079334}
}

@article{anastasiev2022stroke,
  title = {Supervised Myoelectrical Hand Gesture Recognition in Post-Acute Stroke Patients with Upper Limb Paresis on Affected and Non-Affected Sides},
  author = {Anastasiev, Alexey and Kadone, Hideki and Marushima, Aiki and Watanabe, Hiroki and Zaboronok, Alexander and Watanabe, Shinya and Matsumura, Akira and Suzuki, Kenji and Matsumaru, Yuji and Ishikawa, Eiichi},
  journal = {Sensors},
  volume = {22},
  number = {22},
  pages = {8733},
  year = {2022},
  doi = {10.3390/s22228733}
}

@article{bao2024poststroke,
  title = {Deep Learning Based Post-Stroke Myoelectric Gesture Recognition: From Feature Construction to Network Design},
  author = {Bao, Tianzhe and Lu, Zhiyuan and Zhou, Ping},
  journal = {IEEE Transactions on Neural Systems and Rehabilitation Engineering},
  year = {2024},
  note = {Early access},
  doi = {10.1109/TNSRE.2024.3521583}
}

@article{hochreiter1997lstm,
  title = {Long Short-Term Memory},
  author = {Hochreiter, Sepp and Schmidhuber, J{\"u}rgen},
  journal = {Neural Computation},
  volume = {9},
  number = {8},
  pages = {1735--1780},
  year = {1997}
}

@inproceedings{kipf2017gcn,
  title = {Semi-Supervised Classification with Graph Convolutional Networks},
  author = {Kipf, Thomas N. and Welling, Max},
  booktitle = {International Conference on Learning Representations},
  year = {2017}
}

@inproceedings{hamilton2017graphsage,
  title = {Inductive Representation Learning on Large Graphs},
  author = {Hamilton, William L. and Ying, Rex and Leskovec, Jure},
  booktitle = {Advances in Neural Information Processing Systems},
  year = {2017}
}

@inproceedings{velickovic2018gat,
  title = {Graph Attention Networks},
  author = {Veli{\v{c}}kovi{\'c}, Petar and Cucurull, Guillem and Casanova, Arantxa and Romero, Adriana and Li{\`o}, Pietro and Bengio, Yoshua},
  booktitle = {International Conference on Learning Representations},
  year = {2018}
}

@article{hinton2015distilling,
  title = {Distilling the Knowledge in a Neural Network},
  author = {Hinton, Geoffrey and Vinyals, Oriol and Dean, Jeff},
  journal = {arXiv preprint arXiv:1503.02531},
  year = {2015}
}

@article{sri2021dnn,
  title = {Classification of 41 Hand and Wrist Movements via Surface Electromyogram Using Deep Neural Network},
  author = {Sri-Iesaranusorn, Panyawut and Chaiyaroj, Attawit and Buekban, Chatchai and Dumnin, Songphon and Pongthornseri, Ronachai and Thanawattano, Chusak and Surangsrirat, Decho},
  journal = {Frontiers in Bioengineering and Biotechnology},
  volume = {9},
  pages = {548357},
  year = {2021},
  doi = {10.3389/fbioe.2021.548357}
}

@article{li2025transfer,
  title = {Deep End-to-End Transfer Learning for Robust Inter-Subject and Inter-Day Hand Gesture Recognition Using Surface EMG},
  author = {Li, Jianfeng and Jiang, Xinyu and Fan, Jiahao and Geng, Yanjuan and Jia, Fumin and Dai, Chenyun},
  journal = {Biomedical Signal Processing and Control},
  volume = {100},
  pages = {106892},
  year = {2025},
  doi = {10.1016/j.bspc.2024.106892}
}

@article{mohammadiazni2026transfer,
  title = {Hand Gesture Intention Detection Using sEMG and Transfer Learning in Stroke Survivors},
  author = {Mohammadiazni, M. and Huszar, K. and Peters, S. and Trejos, A. L.},
  journal = {IEEE Journal of Biomedical and Health Informatics},
  year = {2026},
  note = {Early access, published May 13, 2026},
  doi = {10.1109/JBHI.2026.3693109}
}

@inproceedings{akiba2019optuna,
  title = {Optuna: A Next-generation Hyperparameter Optimization Framework},
  author = {Akiba, Takuya and Sano, Shotaro and Yanase, Toshihiko and Ohta, Takeru and Koyama, Masanori},
  booktitle = {Proceedings of the 25th ACM SIGKDD International Conference on Knowledge Discovery and Data Mining},
  year = {2019}
}

@article{welch1967fft,
  title = {The Use of Fast Fourier Transform for the Estimation of Power Spectra: A Method Based on Time Averaging Over Short, Modified Periodograms},
  author = {Welch, Peter D.},
  journal = {IEEE Transactions on Audio and Electroacoustics},
  volume = {15},
  number = {2},
  pages = {70--73},
  year = {1967}
}

\end{document}